\documentclass{article}

\usepackage{arxiv}

\usepackage[utf8]{inputenc} 
\usepackage[T1]{fontenc}    
\usepackage{hyperref}       
\usepackage{url}            
\usepackage{booktabs}       
\usepackage{amsfonts}       
\usepackage{nicefrac}       
\usepackage{microtype}      
\usepackage{lipsum}
\usepackage{comment}
\usepackage{graphicx}

\title{CAT STREET: Chronicle Archive of Tokyo Street-fashion}

\author{
  Satoshi Takahashi\thanks{We would like to thank Editage (www.editage.com) for English language editing. This work was supported by the Japan Society for the Promotion of Science (JSPS) KAKENHI [grant number 20K02153].}\\
  College of Science and Engineering\\
  Kanto Gakuin University,\\
  1-50-1 Mutsuura, Kanazawa, Yokohama, Kanagawa, JAPAN\\
  \texttt{satotaka@kanto-gakuin.ac.jp} \\
  \And
  Keiko Yamaguchi\\
  Graduate School of Economics\\
  Nagoya University\\
  Furocho, Chikusa, Nagoya, Aichi, JAPAN\\
  \texttt{keiko.yamaguchi@soec.nagoya-u.ac.jp} \\
  \And
  Asuka Watanabe\\
  Department of the Science of Living\\
  Kyoritsu Women's Junior College\\
  2-6-1 Hitotsubashi, Chiyoda, Tokyo, JAPAN\\
  \texttt{asuka\_watanabe@kyoritsu-wu.ac.jp} \\
}

\begin{document}
\maketitle

\begin{abstract}
The analysis of daily-life fashion trends can provide us a profound understanding of our societies and cultures. However, no appropriate digital archive exists that includes images illustrating what people wore in their daily lives over an extended period. In this study, we propose a new fashion image archive, Chronicle Archive of Tokyo Street-fashion (CAT STREET), to shed light on daily-life fashion trends. CAT STREET includes images showing what people wore in their daily lives during 1970--2017, and these images contain timestamps and street location annotations. This novel database combined with machine learning enables us to observe daily-life fashion trends over a long term and analyze them quantitatively. To evaluate the potential of our proposed approach with the novel database, we corroborated the rules-of-thumb of two fashion trend phenomena that have been observed and discussed qualitatively in previous studies. Through these empirical analyses, we verified that our approach to quantify fashion trends can help in exploring unsolved research questions. We also demonstrate CAT STREET's potential to find new standpoints to promote the understanding of societies and cultures through fashion embedded in consumers' daily lives.
\end{abstract}


\section{Introduction}
In fashion research, the advantages of digital fashion archives have continued to attract attention not only for business improvement but also for understanding our societies from a cultural perspective. In particular, the fashion style adopted in daily life is an important aspect of culture. As noted by \cite{lancioni1973brief}, fashion is 'a reflection of a society's goals and aspirations'; people choose fashion styles within the social contexts in which they are embedded. The analysis of daily-life fashion trends can provide an in-depth understanding of our societies and cultures.

Kroeber \cite{kroeber1919principle} analyzed fashion trends manually and measured features of women's full evening toilette, e.g. skirt length and skirt width, which were collected from fashion magazines from 1844 to 1919. After this work, many researchers conducted similar studies \cite{lowe1993quantitative}. Belleau \cite{belleau1987cyclical} examined the skirt length, waist emphasis, and silhouette of women's day dresses from 1860 to 1980. Lowe and Lowe \cite{lowe1990velocity} measured features of women's formal evening dress, e.g. skirt length and skirt width, from 1789 to 1980. Robenstine and Kelley \cite{robenstine1981relating} examined silhouettes of male and female clothes in portraits and fashion illustrations between 1715 and 1914. Their work surveyed old fashion magazines, portraits, and illustrations and focused mainly on formal fashion, rather than daily-life fashion.

The analysis of daily-life fashion trends has two major issues. The first issue is the absence of an appropriate daily-fashion image archive. Many researchers, museums, and research institutions have created their own fashion image databases. However, the fashion databases proposed in previous studies have several limitations in terms of their approach to daily-life fashion trends; they cannot be used to analyze how the daily-life clothing styles of people change and what drives these changes.

The second issue is that analyzing fashion trends is labor-intensive. Kroeber \cite{kroeber1919principle} is one of the representative works in the early stage of quantitative analysis of fashion trends. This type of quantitative analysis needs a large amount of human resources to select appropriate images, classify images, and measure features with a ruler. This labor-intensive approach has been applied in this field for a long time. Furthermore, manually managing large modern fashion image archives is cumbersome. To solve this issue, we utilize machine learning (ML). ML is a computer algorithm that learns procedures based on sample data, e.g. human task results, and imitates the procedures. In recent years, ML contributed to the development of digital humanities. For instance, Kestemont et al. \cite{kestemont2017lemmatization} applied ML to the lemmatization of Middle Dutch datasets. Wevers and Smits \cite{wevers2020visual} demonstrated that ML can separate photographs from illustrations and classify the data based on visually similar advertisements in an archive of digitized Dutch newspapers. Lang and Ommer \cite{lang2018attesting} used ML to retrieve individual hand gestures from illustrations in Codex Manesse, poetry written in Middle High German in the fourteenth century. In the fashion field, modern fashion styles are complex and diverse. Applying ML methodologies to fashion image archives helps us quantify fashion trends more precisely and efficiently.

In this study, we shed light on research questions about daily-life fashion trends using multiple digital archives and deep learning (DL), which is one of the mainstream methods in ML. To quantify daily-life fashion trends, we built CAT STREET, a new fashion image archive consisting of long-term fashion images that reflect what women wore in their daily lives from 1970 to 2017 in Tokyo with street location annotations. We corroborated the rules-of-thumb of two fashion trend phenomena, namely how economic conditions relate to fashion style share in a long time span and how fashion styles emerge in the street. Through the empirical analyses of these phenomena, we demonstrate that our database and approach have the potential to promote our understanding of societies and cultures.

\section{Related Work}
\subsection{Deep Learning as a Machine Learning Method}
\begin{figure*}[htbp]
   \begin{center}
     \includegraphics[width=0.93\linewidth]{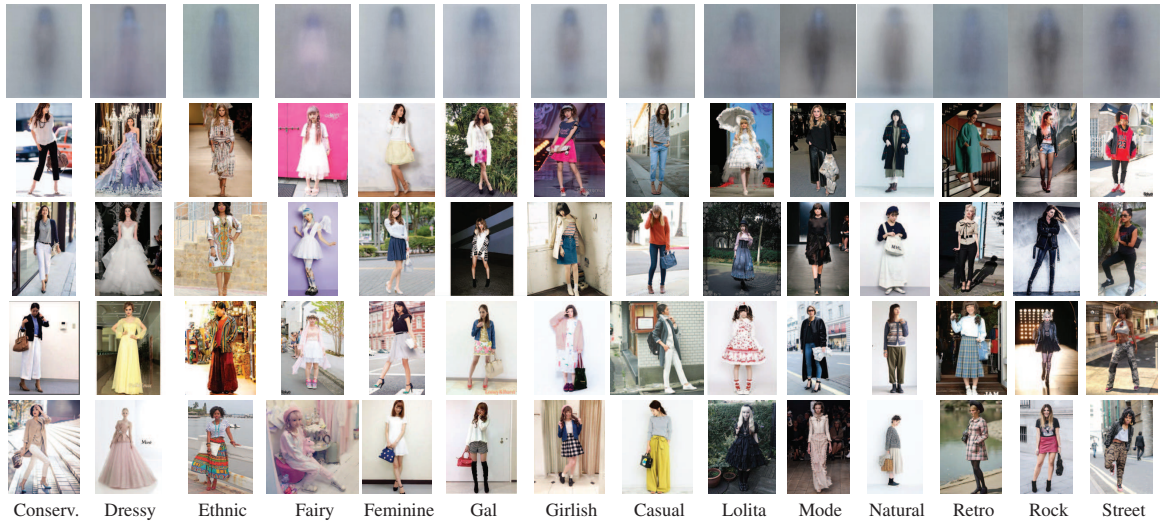}
   \end{center}
   \caption{Examples of the classification of photographic images into fashion style categories \cite{takagi2017makes}}
   \label{fig:fig1}
\end{figure*}

\begin{figure*}[htbp]
   \begin{center}
     \includegraphics[width=0.93\linewidth]{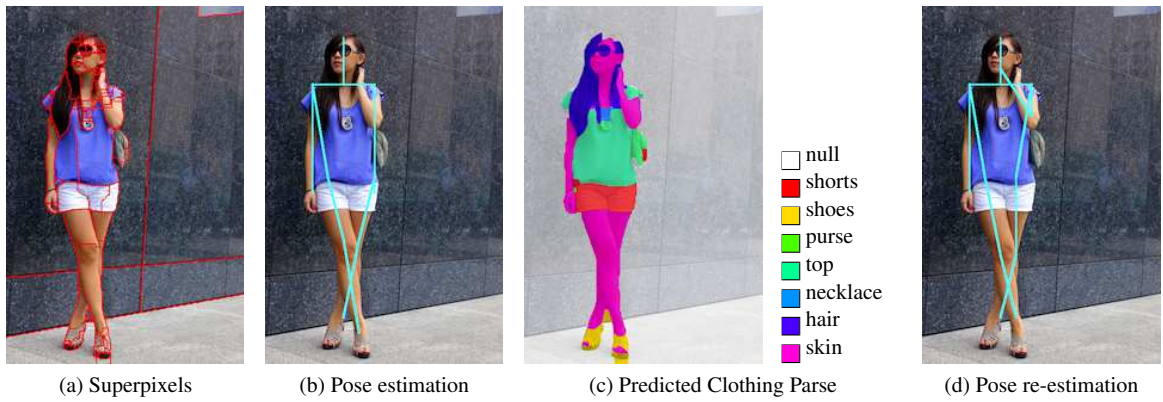}
   \end{center}
   \caption{Example of object recognition and segmentation \cite{yamaguchi2012parsing}}
   \label{fig:fig2}
\end{figure*}

\begin{figure*}[htbp]
   \begin{center}
     \includegraphics[width=0.5\linewidth]{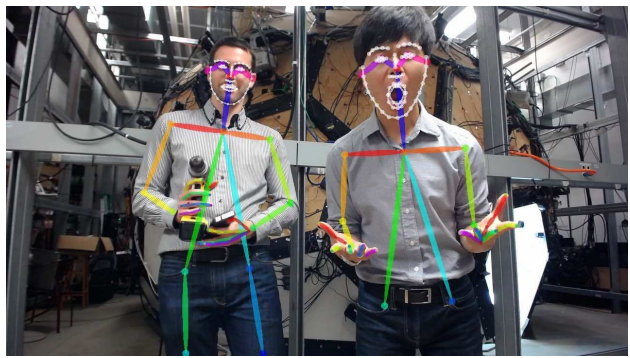}
   \end{center}
   \caption{Example of the detection of skeletal structures \cite{Zhe2019openpose}}
   \label{fig:fig3}
\end{figure*}

ML can be used for tasks such as classification, object recognition, and segmentation. Classification is a task to classify input data by assigning it to one or more categories. In fashion, some models were proposed to classify photographic images into fashion style categories (Fig. \ref{fig:fig1}) \cite{takagi2017makes, sun2015clothing}. Object recognition and segmentation are tasks to detect objects in the input data. They can be used to detect parts of clothes or poses of subjects (Figs. \ref{fig:fig2} and \ref{fig:fig3}) \cite{Zhe2019openpose, yamaguchi2012parsing, bulat2020toward, zhang2020distribution, martinsson2019semantic, zhang2019brief, kuang2020deep, seo2019hierarchical}. By combining these tasks, ML can select appropriate images, classify the images into categories, and measure the features on behalf of a human to improve the efficiency of analyzing digital archives.

\begin{table*}[b]
   \begin{center}
      \begin{tabular}{ccccc}
         \hline
         Database name                           &Number of images  & Geographical information & Time stamp                                          & Fashion style \\
         \hline \hline
         Fashionista \cite{yamaguchi2012parsing}          &158,235           &                          &                                                     &No\\
         Hipster Wars \cite{Kiapour2014}           &1,893             &                          &                                                     &Yes\\
         DeepFashion \cite{liu2016deepfashion}                &800,000           &	                       &                                                     &Yes\\
         Fashion 144k \cite{simo2015neuroaesthetics}	       &144,169	        &City unit                 &		                                               &Yes\\
         FashionStyle14 \cite{takagi2017makes}          &13,126	           &		                    &                                                     &Yes\\
         When Was That Made? \cite{vittayakorn2017made}&100,000           &		                    &\begin{tabular}{c}1900–2009\\Decade unit\end{tabular}&No\\
         Fashion Culture Database \cite{abe2018fashion}   &76,532,219        &City unit                 &\begin{tabular}{c}2000–2015\\Date unit\end{tabular}  &No\\
         CAT STREET (Our Database)	             &14,679	           &Street unit               &\begin{tabular}{c}1970–2017\\Date unit\end{tabular}  &No\\
         \hline
      \end{tabular}
   \end{center}
   \caption{Popular fashion databases used in computer science}
   \label{table:table1}
\end{table*}

\begin{figure*}[h]
   \begin{center}
     \includegraphics[width=0.5\linewidth]{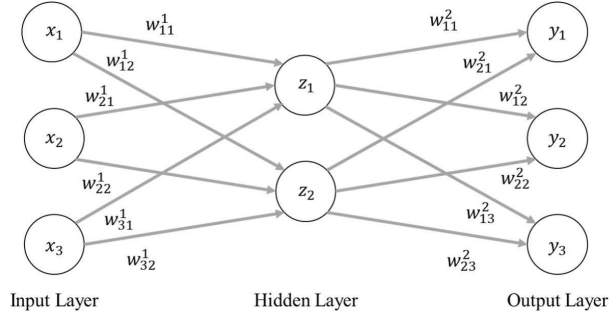}
   \end{center}
   \caption{Example of a neural network}
   \label{fig:fig4}
\end{figure*}

DL is currently one of the mainstream methods in ML \cite{lecun2015deep, deng2014deep}and handles ML tasks with high accuracy. DL is based on neural network technology. Neural networks are inspired by the biological neural networks of animal brains and consist of nodes and edges. Nodes are connected to each other by edges, and each edge has a weight. Fig. \ref{fig:fig4} shows an example of a neural network that predicts the weather for the next day. In the figure, $x_{i}$, $y_{i}$, and $z_{i}$ are nodes, and $w_{jk}^i$ represents the weight. The input data, $x_{1}$, $x_{2}$, and $x_{3}$, correspond to today's air temperature, humidity, and weather, respectively. The output data correspond to the probability of the next day's weather: $y_{1}$ is sunny, $y_{2}$ is cloudy, and $y_{3}$ is rain. This model is called a three-layer neural network: $x_{i}$ represents the input layer, $z_{i}$ represents the hidden layer, and $y_{i}$ represents the output layer. $z_{1}$ is calculated from the inputs $x_{1}$, $x_{2}$, and $x_{3}$ and the weights $w_{11}^1$, $w_{21}^1$, and $w_{31}^1$ by using a function, e.g. a sigmoid function. The other outputs, $y_{i}$ and $z_{i}$, are calculated in a similar manner. Before using a neural network, it must be trained to estimate the appropriate weights, $w_{jk}^i$, by using weather information from the past. This step is called training, and the weather information used for the training is called the training data/dataset.

\begin{figure*}[htbp]
   \begin{center}
     \includegraphics[width=0.93\linewidth]{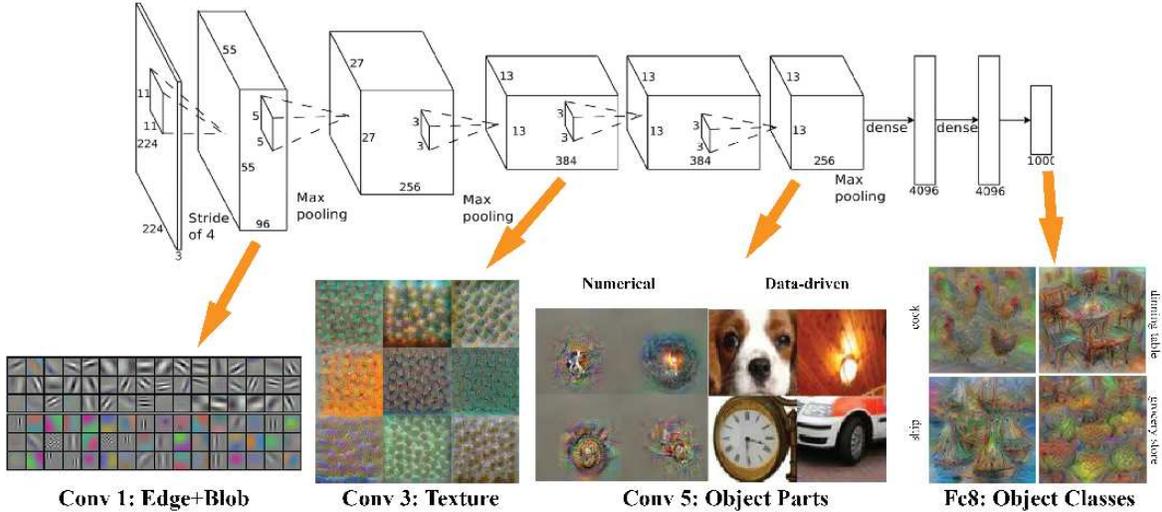}
   \end{center}
   \caption{Example of convolutional layers and pooling layers \cite{mahendran2015understanding}}
   \label{fig:fig5}
\end{figure*}

When a neural network is applied to complicated tasks, e.g. image classification and object recognition in images, many inputs, outputs, and deeper hidden layers may be required. This type of network is called a deep neural network (DNN), and the application of this model is called DL \cite{lecun2015deep, deng2014deep}. In the last few decades, dramatic advances have been achieved in DNN, especially in image recognition fields, owing to the development of the convolutional neural network (CNN) \cite{schmidhuber2015deep}. CNN is a technique that compacts information and is composed of convolution layers and pooling layers (Fig. \ref{fig:fig5}) \cite{mahendran2015understanding}. The characteristic idea of CNN is that of a filtering technique. CNN slices input image data using a filter, e.g. a 3 × 3 filter, and performs calculations with the sliced data. Through this procedure, the input data are condensed in stages to extract noteworthy information. The first hidden layers react to simple features such as edges and blobs, the middle layers react to patterns and texture, and the last layers react to object parts. The output layer considers the combinations of object parts and classifies the input data into object classes \cite{mahendran2015understanding}. In recent years, to improve the capability of DNN and its scope of application, many techniques and network structures have been proposed, such as the recurrent neural network (RNN), long-short term memory (LSTM), and attention \cite{Vaswani2017}, for speech and audio processing, natural language processing, information retrieval, object recognition, segmentation, and classification \cite{deng2014deep}.

Fashion styles are complex combinations of clothing attributes, and DNN can consider attributes ranging from simple patterns, e.g. stripes, dots, and lattices, to combinations of objects, e.g. skirts, pants, shirts, and coats. This characteristic allows DNN to recognize fashion styles. Takagi et al. \cite{takagi2017makes} showed that DNN models can classify fashion styles and recognize the differences between fashion styles. DNN can even distinguish between the Fairy and Lolita styles, which are difficult for fashion non-experts to classify as they have similar features such as excessively frilly dresses. To analyze daily fashion composed of multiple features and tastes, we applied DNN to perform the necessary tasks to extract trends from fashion images.

\subsection{Digital Fashion Archives}
In recent years, digital fashion archives have been proactively built in the fields of museology and computer science. However, the construction of these archives was prompted by different research objectives. Many museums and research institutions have digitized their fashion collections \cite{Vogue, JapaneseFashionArchive, BostonMuseum, MET, EuropeanaFashion, TheMuseumatFIT, LACMAFashion, UNT} to conserve these historically valuable collections and make them available to the public. Vogue digitized their magazines from 1892 to date, the Japanese Fashion Archive collected representative Japanese fashion items from each era, and the Boston Museum of Fine Arts archived historical fashion items from all over the world.

However, in computer science, the motivation to build digital fashion archives is generally short-sighted and primarily to predict the next fashion trends that can be used by vendors to plan production or by online stores to improve recommendation engines. In contrast, in this study the aim is to describe the long-term fashion trend and explain why it occurred. Several studies have created their own fashion image databases to fit the fashion business issues they focused on \cite{yamaguchi2012parsing, liu2016deepfashion, simo2015neuroaesthetics, takagi2017makes, abe2018fashion, Kiapour2014}. Table \ref{table:table1} presents the best-known public databases from fashion studies in computer science. Fashionista is a representative fashion image database from the early stages of fashion analysis that uses computer vision techniques \cite{yamaguchi2012parsing}. The images in Fashionista were collected from Chictopia.com, a fashion blog. They contain pose annotations for fourteen body parts and clothing labels on superpixel regions. There are no location, timestamp, or fashion style annotations.

Subsequently, many fashion image databases with auxiliary information have been created, such as Hipster Wars \cite{Kiapour2014}, DeepFashion \cite{liu2016deepfashion}, Fashion 144k \cite{simo2015neuroaesthetics}, FashionStyle14 \cite{takagi2017makes}, When Was That Made? \cite{vittayakorn2017made}, and Fashion Culture Database \cite{abe2018fashion}. Hipster Wars, DeepFashion, and FashionStyle14 have fashion style labels as training data for fashion style classification. These fashion images were collected from Google Image Search \cite{Kiapour2014}, online shopping websites \cite{liu2016deepfashion}, and Internet crawlers \cite{simo2015neuroaesthetics}. Furthermore, Fashion 144k, When Was That Made?, and Fashion Culture Database have location and/or timestamp annotations. Fashion 144k has city-level location labels \cite{takagi2017makes}; When Was That Made? has decade labels such as 90s, 1954–1957, and 1920s \cite{vittayakorn2017made}; and Fashion Culture Database has city-level location labels and timestamps \cite{abe2018fashion}.

Unfortunately, the fashion databases proposed in previous studies have several limitations in terms of their approach to daily-life fashion trends. First, the level of detail of location annotation in existing databases is insufficient to analyze daily-life fashion trends. Fashion is a way to express oneself and is affected by a wide variety of social factors, such as culture, economic conditions, historical events, and social position \cite{lancioni1973brief}. People belong to a social community, and some social communities have their own distinct fashion styles. In addition, social communities have their own territory. For instance, 'Shibuya Fashion' is a fashion style for young ladies that originated from a famous fashion mall in Shibuya \cite{kawamura2006japanese}. Young ladies dressed in 'Shibuya Fashion' frequent Shibuya, one of the most famous fashion-conscious streets in Japan. Hence, we need fashion image data with location annotations at the street level to focus on what people wear in their daily lives.

Second, most databases consist of recent fashion images from the last decade because they were obtained from the Internet. The periods covered by these databases might not be sufficiently long to determine fashion trends over longer periods. By examining how fashion changes over extended periods, sociologists and anthropologists have found that fashion has decadal-to-centennial trends and cyclic patterns \cite{kroeber1919principle}. An example is the hemline index, which is a well-known hypothesis that describes the cyclic pattern in which skirt lengths decrease when economic conditions improve \cite{Dhanorkar2015}. This pattern was determined based on observations of skirt lengths in the 1920s and 30s. 

As an exception to fashion databases in computer science, Vittayakorn, Berg, and Berg \cite{vittayakorn2017made} collected fashion images from 1900 to 2009 to investigate how vintage fashion influenced fashion show collections in the period 2000–2014. However, these images, labeled according to the decade, do not seem to capture the daily-fashion styles of people. Furthermore, we need data on how and what consumers wear to express their fashion styles. However, fashion photographs on the Internet are one of two types: one displays clothes that consumers themselves choose to wear, and the other consists of photographs taken by professional photographers to promote a clothing line. As previous studies built databases by collecting fashion images from the Internet, existing databases do not reflect only the fashion styles chosen by consumers. This is the third limitation of existing fashion data archives.

From the above previous findings and limitations, it can be inferred that we need a fashion database that has images with more granular date labels and covers a long period of at least a few decades to determine trends and cyclic patterns. Taken together, no existing database has images that span over a long period with both timestamp and location information. In this study, we define daily-life fashion trends as trends of fashion styles that consumers adopt in their daily lives. Daily-life fashion exists on the streets \cite{kawamura2006japanese, Rocamora2008}, changes over time, and trends and cyclic patterns can be observed over extended periods. Our proposed database, CAT STREET, is an attempt to solve some of these limitations.

\section{CAT STREET: Chronicle Archive of Tokyo Street-fashion}
We created CAT STREET via the following steps. We took street-fashion photographs once or twice a month in fashion-conscious streets such as Harajuku and Shibuya in Tokyo from 1980 to date. In addition, we used fashion photographs from a third-party organization taken in the 1970s at monthly intervals in the same fashion-conscious streets. Next, by using images from the two data sources, we built a primary image database that has timestamps from 1970 to 2017. The photographs from the third-party organization did not have location information; hence, we could only annotate the fashion images taken since 1980 with street tags.

Fashion styles are different for men and women. To focus on women's fashion trends in CAT STREET, we detected the subject's gender and selected only women's images from the primary image database. Gender detection was performed manually by two researchers, and the detection results were validated reciprocally. Some images from the 1970s are in monochrome; therefore, we gray-scaled all images to align the color tone across all images over the entire period.

\begin{figure*}[htbp]
   \begin{center}
     \includegraphics[width=0.93\linewidth]{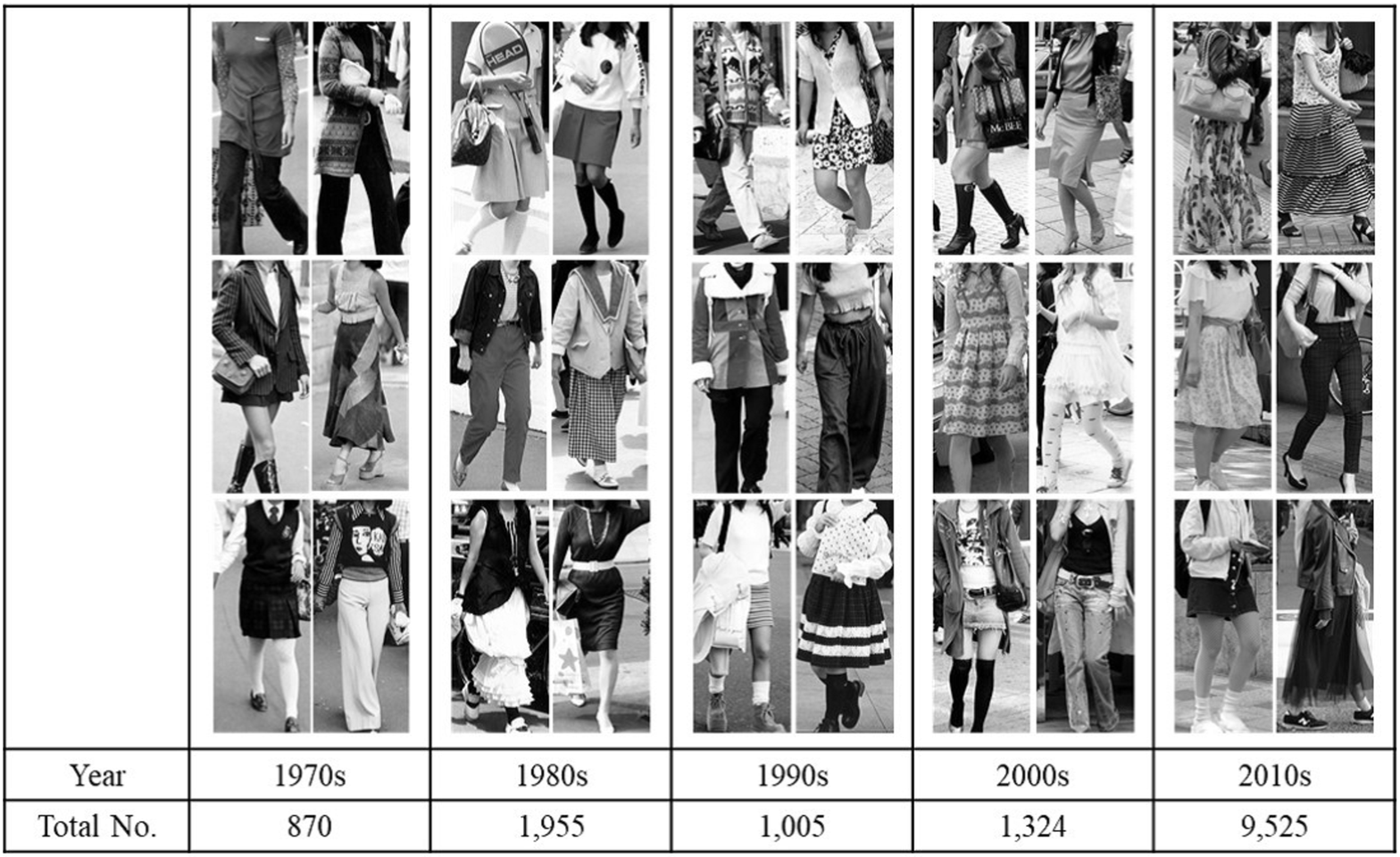}
   \end{center}
   \caption{Data overview of CAT STREET}
   \label{fig:fig6}
\end{figure*}

Street-fashion photographs contain a large amount of noise, which hinders the precise detection of combinations of clothes that people wear. To remove the noise, we performed image pre-processing as the final step in the following manner. We identified human bodies in the photographs using OpenPose \cite{Zhe2019openpose}, an ML algorithm for object recognition, and removed as much of the background image as possible. Subsequently, we trimmed the subject's head to focus on clothing items based on the head position, which was detected using OpenPose. Fig. \ref{fig:fig6} shows an overview of the data contained in CAT STREET. The total number of images in the database is 14,679.

At the end of the database creation process, we checked whether CAT STREET met the requirements for a database to capture daily-life fashion trends. First, CAT STREET comprises street-snap photographs in fashion-conscious streets in Tokyo. It reflects the fashion styles women choose in their real lives and does not include commercial fashion images for business purposes. Second, CAT STREET has necessary and sufficient annotations to track fashion trends: monthly timestamps from 1970 to 2017 and street-level location tags. Images in the 1970s do not have street tags; however, they were taken in the same streets as the photographs taken with street tags since 1980. Hence, we could use all the images in CAT STREET to analyze the overall trends of daily-life fashion in fashion-conscious streets as a representative case in Japan. Therefore, we believe that CAT STREET has sufficient features to analyze daily-life fashion trends and the potential to promote the understanding of our societies and cultures through quantitative analysis.

\begin{table*}[tbhp]
   \begin{center}
      \begin{tabular}{ccccc}
         \hline
                           &InceptionResNetV2   &Xception            &ResNet50&VGG19\\
         \hline \hline
         Conservative	   &0.754	            &\bf\underline{0.758}&0.708	&0.620\\
         Dressy	         &\bf\underline{0.940}&0.936	            &0.938	&0.898\\
         Ethnic	         &\bf\underline{0.812}&0.806	            &0.747	&0.668\\
         Fairy	            &\bf\underline{0.901}&0.887	            &0.876	&0.814\\
         Feminine	         &0.724	            &\bf\underline{0.734}&0.681	&0.644\\
         Gal	            &\bf\underline{0.782}&0.779	            &0.757	&0.710\\
         Girlish	         &\bf\underline{0.640}&0.629	            &0.603	&0.523\\
         Casual	         &0.665	            &\bf\underline{0.667}&0.619	&0.563\\
         Lolita	         &\bf\underline{0.949}&0.942	            &0.933	&0.886\\
         Mode	            &0.748	            &\bf\underline{0.754}&0.721	&0.650\\
         Natural	         &\bf\underline{0.793}&0.779    	         &0.710	&0.692\\
         Retro	            &\bf\underline{0.701}&0.700   	         &0.644	&0.590\\
         Rock	            &\bf\underline{0.777}&0.765   	         &0.745	&0.694\\
         Street	         &\bf\underline{0.835}&0.816   	         &0.781	&0.701\\
         \hline
         Weighted Avg.	   &\bf\underline{0.786}&0.781   	&0.747	&0.689\\
         \hline
      \end{tabular}
   \end{center}
   \caption{F1-scores of fine-tuned network architectures. The highest scores are underlined and in boldface}
   \label{table:table2}
\end{table*}

\begin{figure*}[htbp]
   \begin{center}
     \includegraphics[width=0.93\linewidth]{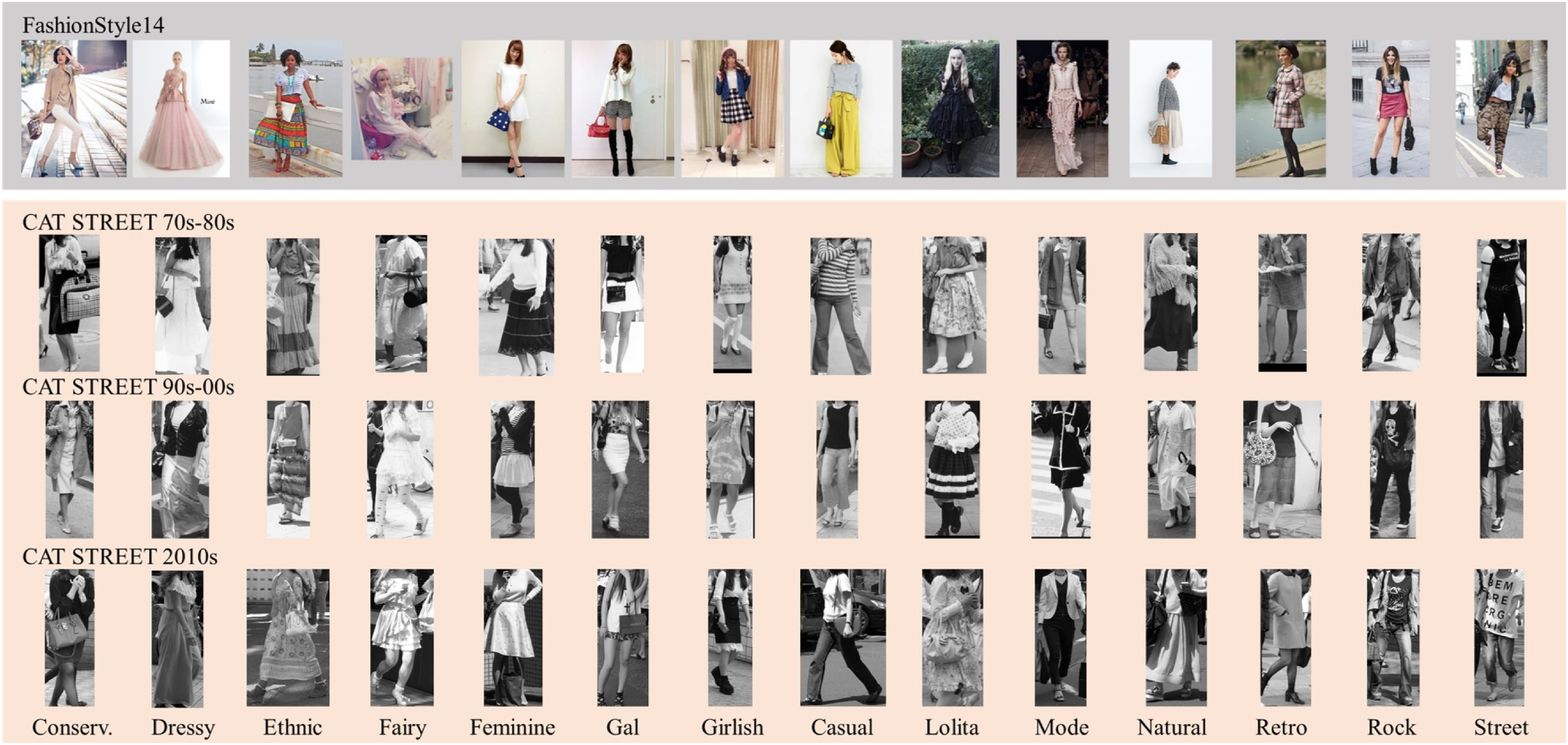}
   \end{center}
   \caption{Overview of FashionStyle14 \cite{takagi2017makes} and sample images in CAT STREET classified into different fashion styles using the fashion style clustering model. 'Conserv.' is an abbreviation for Conservative}
   \label{fig:fig7}
\end{figure*}

\section{Analysis of Daily-life Fashion Trends}
In this section, we report our analysis of the daily-life fashion trends in CAT STREET using DL.

First, we estimated a fashion style clustering model to identify the fashion styles adopted in fashion images. Second, we applied the fashion clustering model to CAT STREET and calculated the extent to which people adopted the fashion styles in each year. Finally, we demonstrated that our approach with CAT STREET has some potential to promote the understanding of our society. To show that our approach enables us to verify fashion phenomena that have been qualitatively discussed previously and propose new standpoints and research questions for expanding fashion trend theories, we corroborated the rules-of-thumb for two fashion trend phenomena in our society, namely how economic conditions relate to fashion style share over a long time and how fashion styles emerge in the street.

\subsection{Fashion Style Clustering Model}
To build a fashion style clustering model, we selected FashionStyle14 \cite{takagi2017makes} as the training dataset. It consists of fourteen style classes, as shown in the first row of Fig. \ref{fig:fig7}. Each class consists of approximately 1,000 images, and the database consists of a total of 13,126 images. The fashion styles of FashionStyle14 were selected by an expert as being representative of modern fashion trends in 2017. By applying the fashion clustering model to CAT STREET, we measured the share of each modern style in each year. We also found the beginnings of the characteristic modern fashion, such as the time when the Fairy style came into fashion. Some limitations of this approach will be discussed later.

We trained four DL network structures as options for our fashion clustering model: InceptionResNetV2 \cite{szegedy2017inception}, Xception \cite{chollet2017xception}, ResNet50 \cite{he2016deep}, and VGG19 \cite{Simonyan2015}. We set weights trained on ImageNet \cite{russakovsky2015imagenet} as the initial weights and fine-tuned them on FashionStyle14 using the stochastic gradient descent algorithm at a learning rate of $10^{4}$. For fine-tuning, we applied k-fold cross-validation with k set as five.

The F1-scores are presented in Table \ref{table:table2}. InceptionResNetV2 yielded the highest F1-scores among the DL network structures for most fashion styles. Its accuracy was 0.787, which is higher than the benchmark accuracy of 0.72 established by ResNet50 trained on FashionStyle14 in the study by Takagi et al. \cite{takagi2017makes}. Therefore, we adopted the DL network structure InceptionResNetV2 as the fashion style clustering model in this study.

\subsection{Fashion Style Clustering Model}
We applied the fashion style clustering model to the images in CAT STREET. Fig. \ref{fig:fig7} shows sample images classified into each fashion style using the fashion style clustering model.

\begin{figure*}[htbp]
   \begin{center}
     \includegraphics[width=0.93\linewidth]{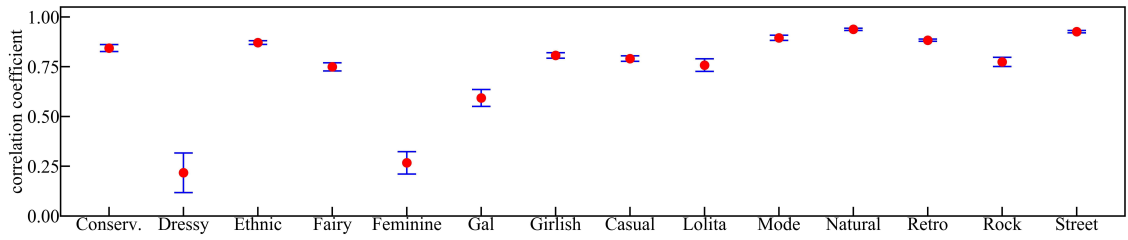}
   \end{center}
   \caption{Average correlation coefficients among five models for fashion styles. The error bars represent the unbiased standard error. 'Conserv.' is an abbreviation for Conservative}
   \label{fig:fig8}
\end{figure*}

\begin{figure*}[htbp]
   \begin{center}
     \includegraphics[width=0.93\linewidth]{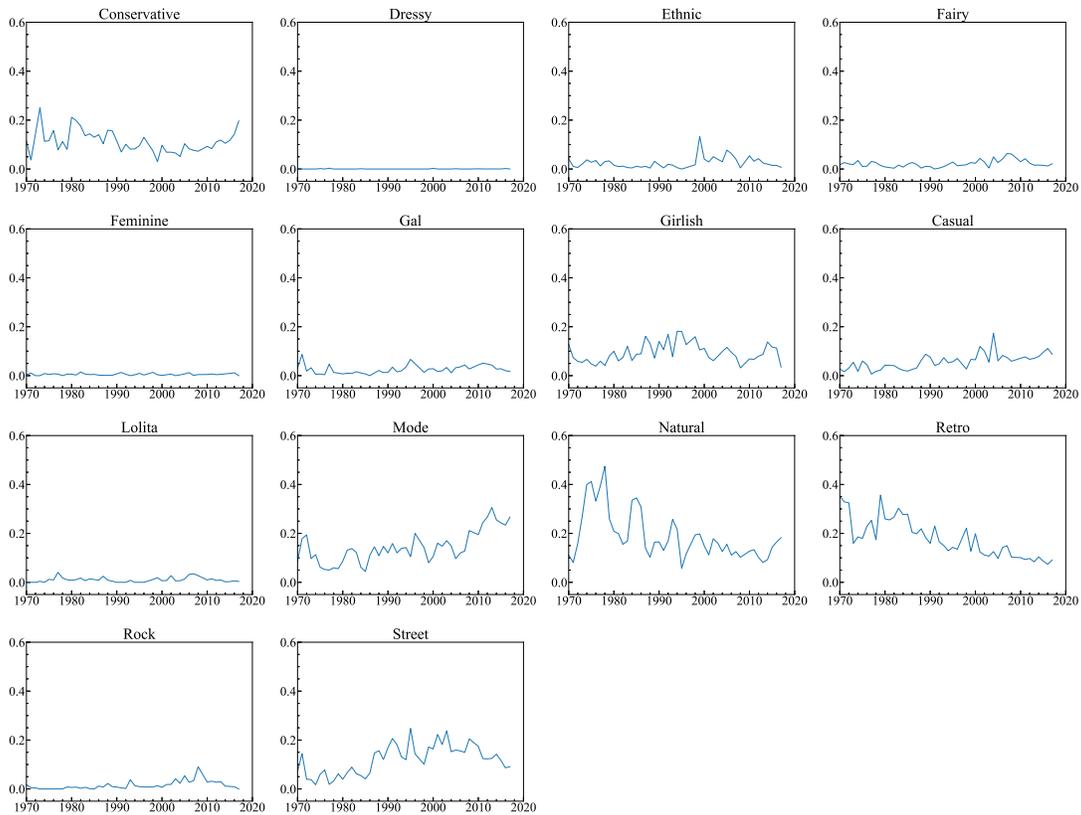}
   \end{center}
   \caption{Style shares of different fashion styles}
   \label{fig:fig9}
\end{figure*}

The fashion style clustering model consisted of five models as we performed five-fold cross-validations when the DL network structure was trained, and each model estimated the style share for each image. To verify the clustering model's robustness in terms of reproducing style shares, we evaluated the time-series correlations among the five models. Fig. \ref{fig:fig8} shows the average correlation coefficients of the five models for fashion styles. Most fashion styles had high correlation coefficients of over 0.8, and the unbiased standard errors were small. Some fashion styles, such as the Dressy and Feminine styles, exhibited low correlations because these styles originally had low style shares. These results indicate that our fashion style clustering model is a robust instrument for reproducing the time-series patterns of style shares. Fig. \ref{fig:fig9} shows the averages of five style shares for the period 1970–2017 for each fashion style. There were no images for 1997 and 2009; we replaced the corresponding zeros with the averages of the adjacent values.

\section{Applications and Discussions}
To evaluate the potential for a quantitative approach to fashion trend analysis with CAT STREET to promote the understanding of our societies, we corroborated two rules-of-thumb for fashion trend phenomena using the estimated style shares in the previous section.

\subsection{How Social Factors Affect Fashion Style Share}
Social contexts prompt people to shape their social identities \cite{tajfel1979integrative} and choose adequate fashion styles to express their identities. Economic conditions are thought to have an impact on people's daily clothes \cite{hemphill2008law}. Occasionally, people aspire for a higher social status and are likely to purchase quality-guaranteed luxury items as a symbol of wealth \cite{nia2000counterfeits}. At the same time, these consumer behaviors rely on the economic conditions in the society.

\begin{figure*}[htbp]
   \begin{center}
     \includegraphics[width=0.93\linewidth]{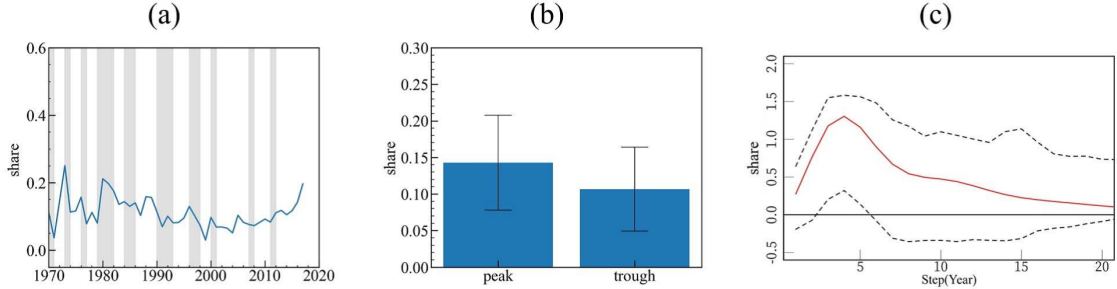}
   \end{center}
   \caption{Conservative style's average style share, ANOVA, and impulse response function. (a) Average style share for the Conservative style. The gray area indicates the recession periods in the Japanese business cycle. (b) ANOVA of the Conservative style share with the economic condition. The error bars represent the standard deviation. (c) Impulse response function for the effect of the GDP growth rate on the averaged time-series Conservative share over time. The dashed lines indicate 95\% confidence limits}
   \label{fig:fig10}
\end{figure*}

To demonstrate the capability of our approach with CAT STREET to verify the above qualitatively discussed relationship between social contexts and the long-term evolution of fashion style share, we investigated how economic conditions relate to the Conservative share, which includes luxury fashion brands and items with that definition, as shown in Fig. \ref{fig:fig10}(a).

With the time-series Conservative share of the five models in CAT STREET as the dependent variable, we performed a one-way analysis of variance (ANOVA) to evaluate the difference in the share level during business-cycle trough/peak periods in the Japanese economy. Fig. \ref{fig:fig10}(b) shows the results obtained. The average Conservative style shares between the business-cycle trough and peak differed significantly ($p < .05$). This result indicates that an economic upturn prompted consumers to choose the Conservative style.

We also applied the vector autoregression (VAR) model to test the Granger causality between the averaged time-series Conservative share and the gross domestic product (GDP) growth rate in Japan. VAR is a statistical model used to express the relationships, that is, Granger causalities, among multiple time-series data, and the Granger causality test is a statistical test to verify that one time-series data point is useful to predict another. In this case, we verified that the GDP growth rate is useful to predict the averaged time-series Conservative share.

To construct the VAR model, we used the averaged time-series Conservative share from 1970 to 2008. The bankruptcy of Lehman Brothers occurred in 2008, and we assumed that it negatively impacted the business of luxury brands \cite{Fujimura} and changed the business structure in the fashion industry. The VAR model with a lagged order of   of the averaged time-series Conservative share and the GDP growth rate is given by the following equation:
$$
y_{t}=\alpha_{y0}+\sum_{n=1}^N\beta_{yn}y_{t-n}+\sum_{n=1}^N\gamma_{yn}x_{t-n}+\epsilon_{yt},
$$

$$
x_{t}=\alpha_{x0}+\sum_{n=1}^N\beta_{xn}y_{t-n}+\sum_{n=1}^N\gamma_{xn}x_{t-n}+\epsilon_{xt}.
$$

The averaged time-series Conservative share at year $t$ is denoted by $y_{t}$ and the GDP growth rate at year $t$ by $x_{t}$; $\epsilon$ represents white noise; and $\alpha$, $\beta$, and $\gamma$ are the parameters of the model and can be estimated using ordinary least squares on each equation.

In terms of Akaike's information criterion (AIC), we selected the VAR model with a lagged order of two ($N = 2$), and the F-test showed that the model was significant ($p < .05$). AIC is an indicator to evaluate a model's fit to data. We also conducted the Granger causality test, which indicated that the GDP growth rate 'Granger-causes' the Conservative share. To verify the effect and significance of the GDP growth rate on the Conservative share, we performed an analysis of the impulse response function (IRF) using the estimated VAR model. IRF analysis is a type of simulation that predicts the change in the Conservative share when the GDP growth rate experiences a one-standard-deviation shock. Fig. \ref{fig:fig10}(c) plots the IRF of the effect of GDP growth rate over time and indicates a significant carry-over effect; the increase in the GDP growth rate prompted consumers to choose the Conservative style after a two-year delay.

The Conservative trend was originally represented by a combination of clothing items that have traditional and old-school aesthetics, and fashion experts often label items from highly recognized, quality-guaranteed luxury brands as the Conservative style \cite{hemphill2008law, nia2000counterfeits}. People who adopt this style exhibit a sense of credibility because the style projects a high social status. The results in this section imply that consumers are more likely to adopt the Conservative style when business conditions are booming, and this definition of style qualitatively justified our results. 

\subsection{How Fashion Styles Boom in the Streets}
There are two representative fashion-conscious streets in Tokyo: Harajuku and Shibuya. Geographically, Harajuku and Shibuya are very close and only one station away from each other; however, they have different cultures. In particular, the daily-life fashion trends in these streets are famously compared to each other. Some qualitative studies pointed out triggers that form the cultural and fashion modes in each street using an observational method \cite{kawamura2006japanese}. However, they simply reported the triggers with respect to the street and style and did not compare how the triggers work on the mode formations in daily-life fashion trends from a macro perspective. This section sheds light on this research gap and explores the potential of our approach with CAT STREET to prompt research questions that can contribute to the expansion of fashion theories.

\begin{figure*}[b]
   \begin{center}
     \includegraphics[width=0.93\linewidth]{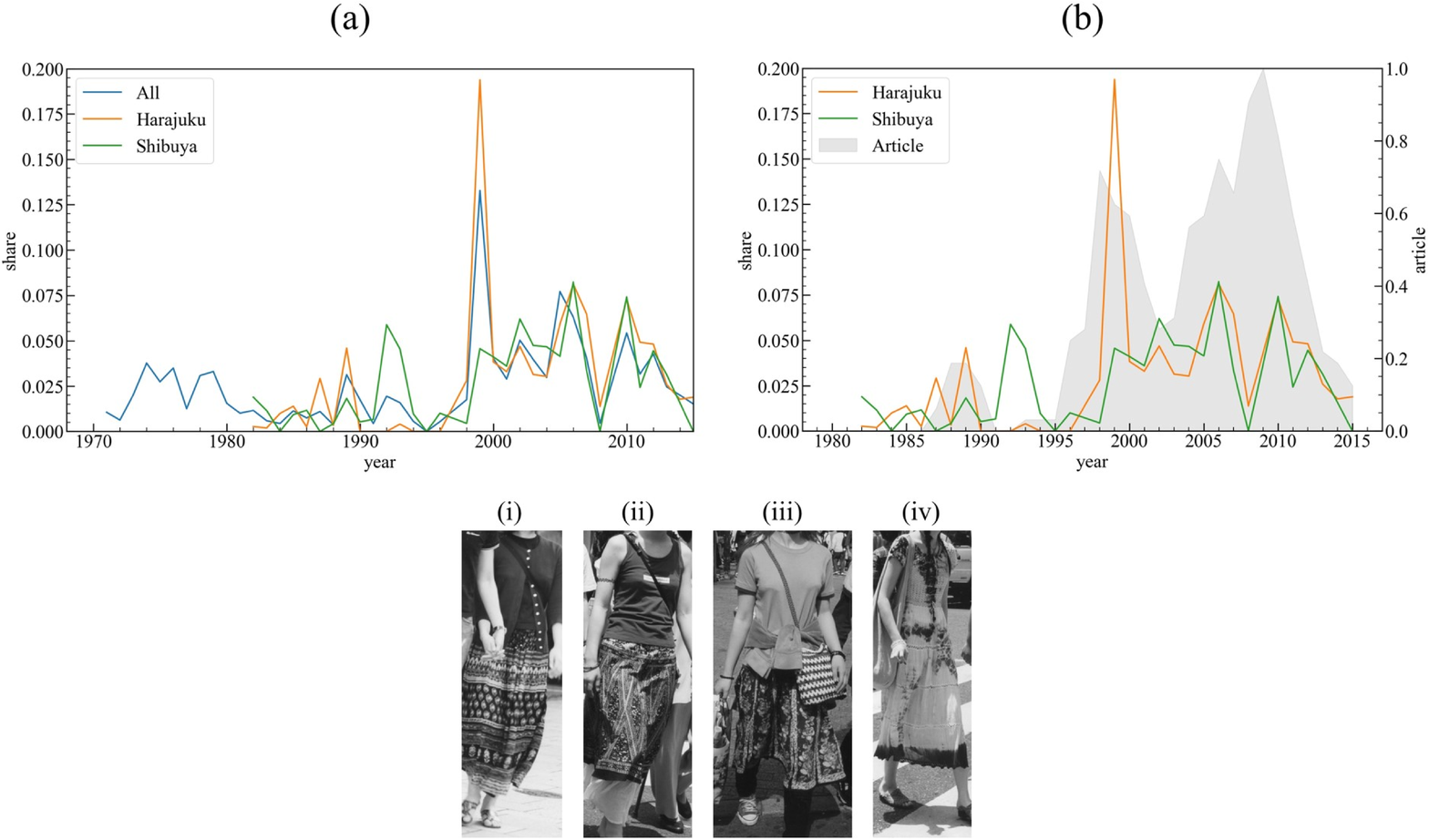}
   \end{center}
   \caption{Ethnic style's simultaneous emergence in both streets. (a) Average Ethnic style share. (b) Number of articles about the Ethnic style. Four images taken in Harajuku (i, ii) and Shibuya (iii, iv) in the late 1990s. The articles were tagged with trend and fashion by Oya Soichi Library. The number of articles was normalized to the range of 0–1}
   \label{fig:fig11}
\end{figure*}

\begin{figure*}[htbp]
   \begin{center}
     \includegraphics[width=0.93\linewidth]{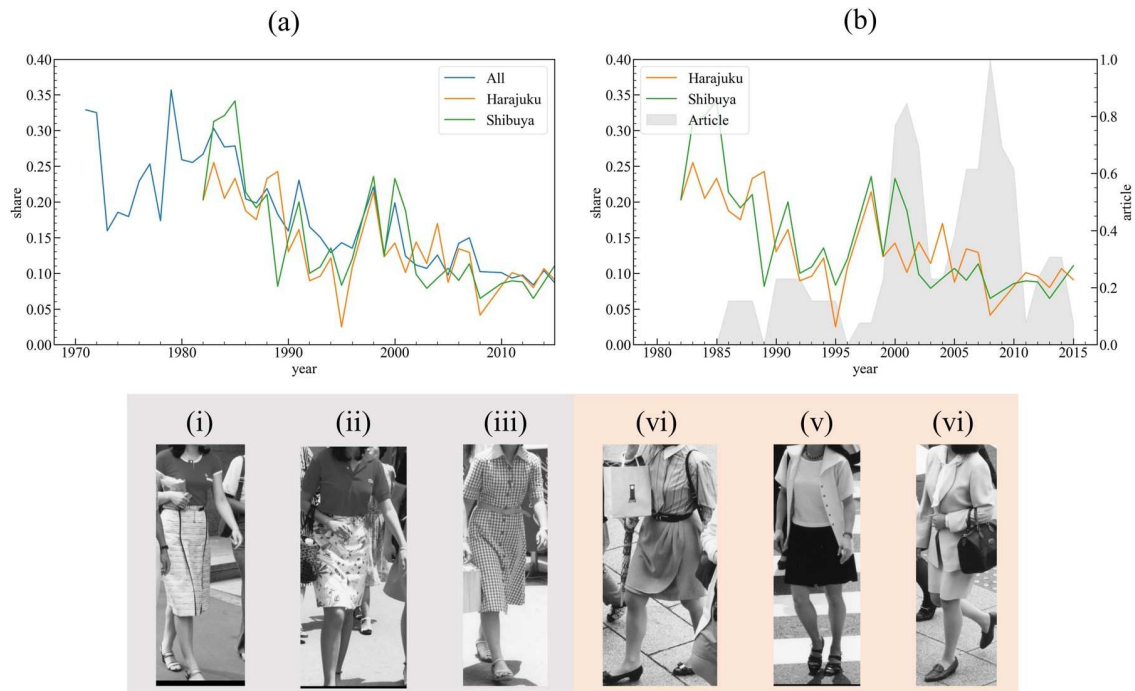}
   \end{center}
   \caption{Retro style's simultaneous emergence in both streets. (a) Average Retro style share. (b) Number of articles about the 80s look. The three images on the left (i–iii) are examples of the Retro style taken in the early 1980s, and the three images on the right (iv–vi) are examples of the Retro style that came into fashion in the late 1990s. The articles were tagged with trend and fashion by Oya Soichi Library. The number of articles was normalized to the range of 0–1}
   \label{fig:fig12}
\end{figure*}

\begin{figure*}[htbp]
   \begin{center}
     \includegraphics[width=0.93\linewidth]{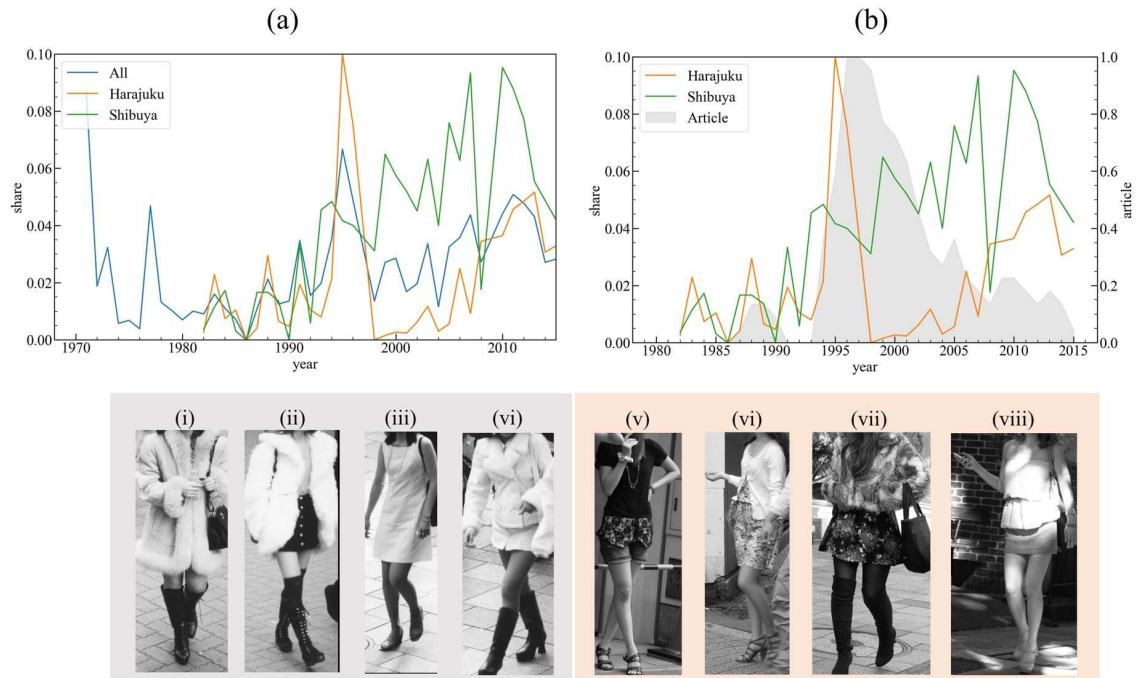}
   \end{center}
   \caption{Gal style's trends in the streets showing different modes. (a) Average Gal style share. (b) Number of articles about the Gal style. The four images on the left were taken in Harajuku (i, ii) and Shibuya (iii, iv) in 1995–1996, and the four images on the right were taken in Harajuku (v, vi) and Shibuya (vii, viii) in the late 2000s. The articles were tagged with trend and fashion by Oya Soichi Library. The number of articles was normalized to the range of 0–1}
   \label{fig:fig13}
\end{figure*}

\begin{figure*}[htbp]
   \begin{center}
     \includegraphics[width=0.93\linewidth]{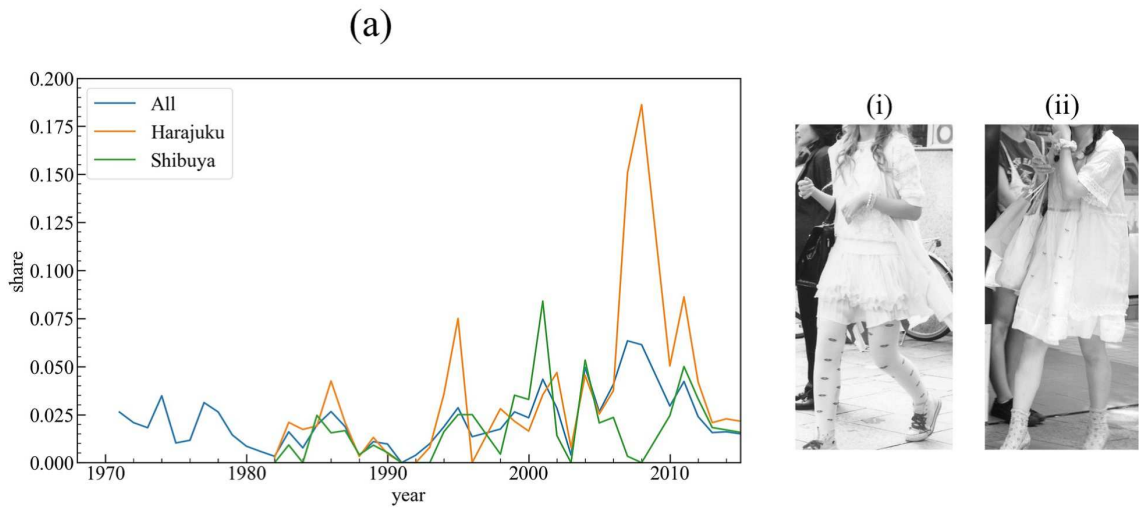}
   \end{center}
   \caption{Fairy style's trends in the streets showing different modes. (a) Average Fairy style share. The two images (i and ii) were taken in Harajuku in the late 2000s}
   \label{fig:fig14}
\end{figure*}

First, we compared two daily-life fashion trends in fashion-conscious streets with CAT STREET to investigate how fashion styles boom and classified the patterns. Fourteen styles were classified into two groups: a group comprising styles that emerged on both streets simultaneously and a group with differing timings for trends observed in the streets. We selected two fashion styles from each group as examples: Ethnic and Retro from the first group and Fairy and Gal from the second group; Figs. \ref{fig:fig11}(a), \ref{fig:fig12}(a), \ref{fig:fig13}(a), and \ref{fig:fig14}(a) show their average style shares, respectively.

\begin{figure*}[htbp]
   \begin{center}
     \includegraphics[width=0.5\linewidth]{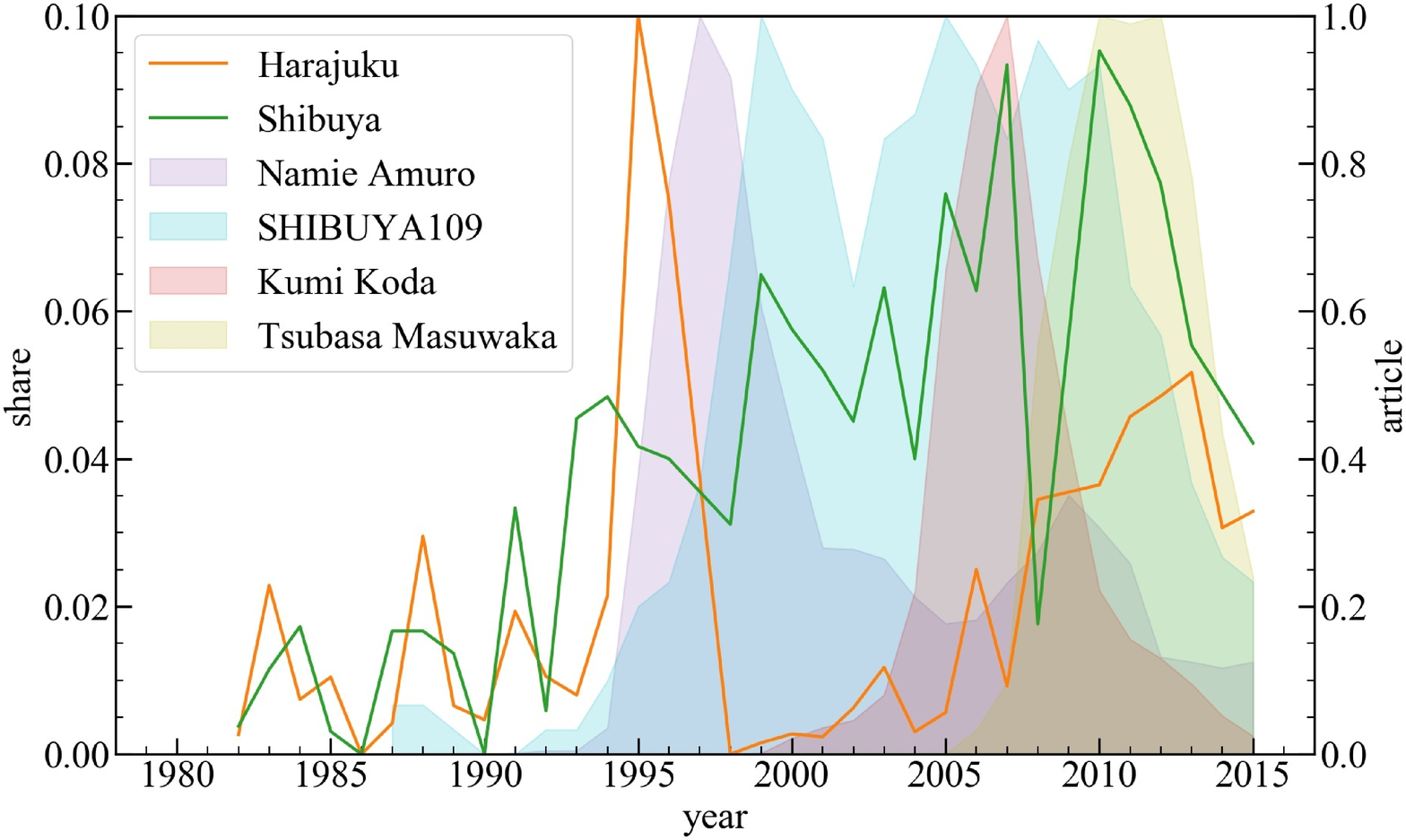}
   \end{center}
   \caption{Gal style's share and 'magazine' trends for its icons. The articles were tagged with trend and fashion by Oya Soichi Library. The number of articles was normalized to the range of 0–1}
   \label{fig:fig15}
\end{figure*}

\begin{figure*}[htbp]
   \begin{center}
     \includegraphics[width=0.5\linewidth]{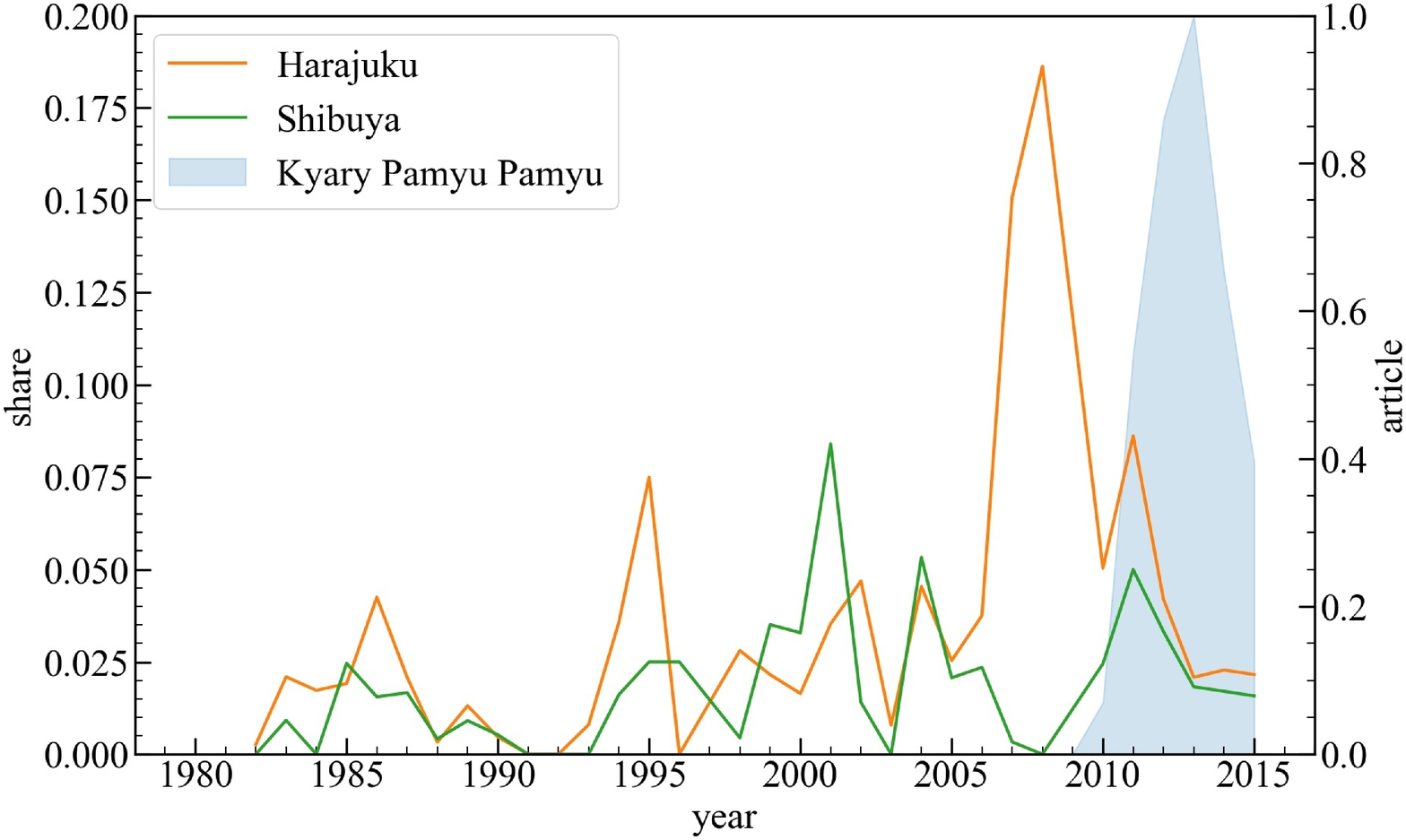}
   \end{center}
   \caption{Fairy style's share and 'magazine' trends for its icon. The articles were tagged with trend and fashion by Oya Soichi Library. The number of articles was normalized to the range of 0–1}
   \label{fig:fig16}
\end{figure*}

\begin{figure*}[htbp]
   \begin{center}
     \includegraphics[width=0.5\linewidth]{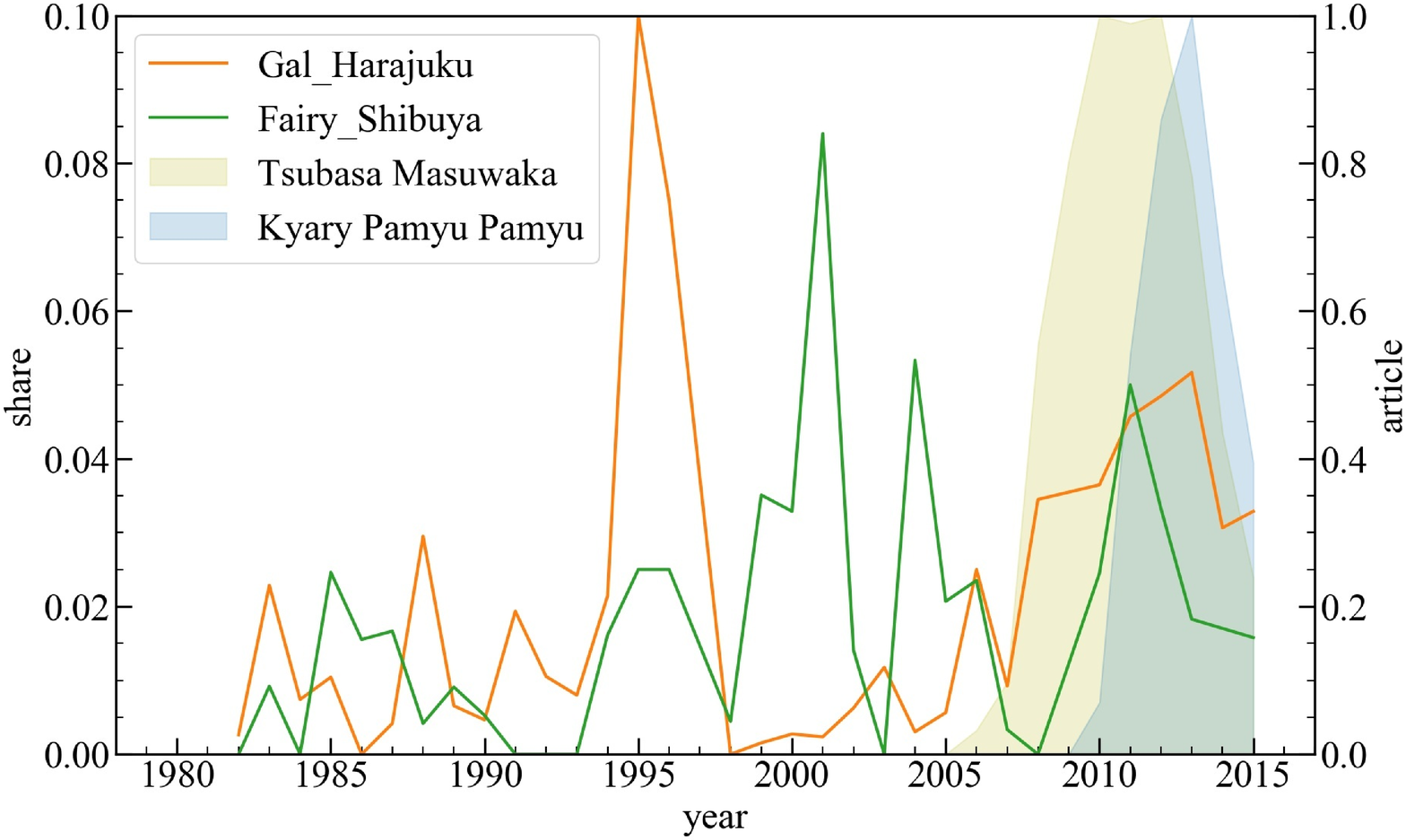}
   \end{center}
   \caption{Gal and Fairy style shares and 'magazine' trends for their icons. The articles were tagged with trend and fashion by Oya Soichi Library. The number of articles was normalized to the range of 0–1}
   \label{fig:fig17}
\end{figure*}

We focused our attention on magazines because fashion brands have built good partnerships with the magazine industry for a long time. Fashion brands regard magazines as essential media to build a bridge between themselves and consumers, and magazines play a role as a dispatcher of fashion in the market. We quantified the 'magazine' trends to capture how magazines or media sent out information to consumers. For this purpose, we used the digital magazine archive of Oya Soichi Library \cite{OYA}. Oya Soichi Library has archived Japanese magazines since the late 1980s and built the digital archive. The digital archive houses about 1,500 magazines, weekly magazines, woman's magazines, and monthly magazines, and one can search for article headlines of about 4.5 million articles. In the fashion and women's lifestyle fields, the archive covers a range of age groups and mass-circulation magazines that people can easily acquire at small bookstores, convenience stores, and kiosks. By searching for headlines including fashion style words and specific topics from articles with tags related to fashion, we could graphically represent the 'magazine' trends that indicate how many articles dealt with the styles and specific topics to spread the information to a mass audience (Fig. \ref{fig:fig11}(b), \ref{fig:fig12}(b), \ref{fig:fig13}(b), \ref{fig:fig15}, \ref{fig:fig16}, and \ref{fig:fig17}).

Finally, to attempt to find new research standpoints about the roles of magazines and media on the formation of fashion trends, we compared the fashion style shares extracted from CAT STREET to the 'magazine' trends in the Oya Soichi Library database in chronological order.

\subsubsection{Simultaneous Emergence of Fashion Styles}
As the first case analysis, we selected two styles, Ethnic and Retro, as examples in the group of styles that simultaneously emerged on both streets. The Ethnic style is inspired by native costumes \cite{bunka1993}. Fig. \ref{fig:fig11}(a) shows the Ethnic style's upward trend in the late 1990s in Harajuku. Simultaneously, the Ethnic style gradually became accepted in Shibuya in the late 1990s and reached the same share level in the mid-2000s. The Retro style, another example in the first group, is an abbreviation of retrospective style \cite{bunka1993, Yoshimura2019}. According to this style's definition, an overall downward trend is plausible in both streets, as shown in Fig. \ref{fig:fig12}(a). Fashion revival is one of the relevant fashion trend phenomena of the Retro style, and there are a wide variety of substyles representing fashion revivals under the Retro style, such as the 60s look, 70s look, and 80s look. In Fig. \ref{fig:fig12}(a), slight peaks can be observed in the early 1980s and late 1990s, suggesting that the 80s look, which was in vogue in the early 1980s, was revived in the late 1990s.

To quantify how magazines spread information about these styles, we plotted the 'magazine' trends by searching for articles with headlines that included terms related to the style names. Figs. \ref{fig:fig11}(b) and \ref{fig:fig12}(b) indicate two relationship patterns between style shares and 'magazine' trends; Fig. \ref{fig:fig11}(b) shows a synchronization phenomenon between style shares and magazine trends, while Fig. \ref{fig:fig12}(b) shows that magazine trends follow share trends.

We interpreted the first pattern as 'mixed'; the shape of 'magazine' trends roughly matches that of style share trends, and this pattern is observed in the Ethnic style. The 'mixed' patterns in the trends suggest that they include two types of phenomena between consumers and magazines. One is that the articles create new trends, spread them, and lead consumers to follow them. The other is that consumers create new style trends by themselves, and articles catch up on the new trends. These two types of phenomena could occur simultaneously or alternately. On the other hand, the second pattern is a 'follow-up'; the articles dealt with the modes that were already in vogue in the streets and contributed to keeping their momentum for a while by spreading the information, and this pattern is observed in the Retro style during the 2000s.

In this analysis, we found two relationship patterns between style shares and 'magazine' trends: 'mixed' and 'follow-up' patterns. We also inferred what types of interactions occur between consumers and magazines in each pattern. However, we could not perform an in-depth analysis on how to distinguish between the two types of consumer–magazine interactions in the case of the 'mixed' pattern because 'magazine' trends generated by searching for article headlines that include terms related to the style names did not always reflect the contents of articles accurately. We also found it difficult to uncover why the article peak in the Retro style during the late 2000s is seemingly irrelevant to the Retro style share. To approach these unsolved research questions as future work, we must perform text mining on article contents and headlines, and categorize the articles into the creating type, which creates and spreads new trends, or the reporting type, which catches up to the new consumer trends; e.g. 'ten trends coming in autumn, check pattern, check pattern, military, big tops, shoes \& boots, foot coordination, fur, Ethnic vs. Nordic, beautiful romper, best of the season' (translated from Japanese by the authors) is categorized as a creating-type article, while 'Ethnic style is a hot topic this spring. We will introduce you to some of the Japanese items that are getting a lot of attention. Crepe, goldfish prints, and more...' (translated from Japanese by the authors) is categorized as a reporting-type article. By measuring the number of articles about the fashion styles and identifying the contents of articles to reflect the types of consumer–magazine relationships, we expect to decompose relationship types in the 'mixed' pattern, interpret seemingly irrelevant relationships, and clarify the role of each relationship type in the trend formation process.

\subsubsection{Emergence of Fashion Styles at Different Times in the Streets}
For the case where fashion trends are observed at different times in the two streets, we focused on two styles: the Gal style, which can be characterized as a slightly sexy homegirl fashion style, and the Fairy style, which comprises the fashion coordination of frilly dresses that reminds people of fairies \cite{Yoshimura2019}.

Figs. \ref{fig:fig13} and \ref{fig:fig14} show how the acceptances of these two styles changed in each street. The Gal style came into fashion around 1995 simultaneously in Harajuku and Shibuya. The style remained in Shibuya, whereas it lost its popularity quickly in Harajuku but re-emerged in the late 2000s. The Fairy style emerged in the late 2000s in Harajuku only and lost its momentum in the early 2010s. As with the first case, we searched 'magazine' trends for these style names and compared them to the style shares. For the Gal style, we found that the style and 'magazine' trends correlated with each other in the mid-80s and the mid-90s. However, magazines gradually lost interest in the style after the mid-00s, and the relationship between the style and 'magazine' trend disappeared accordingly. For the Fairy style, on the other hand, no article in the digital magazine archive included the style name in the headlines. This is because 'Fairy' is a kind of jargon among people who loves the style and magazines would not use it.

What accounts for this difference between the first and the second case? We assumed that the critical factor determining people's choice in fashion, i.e. 'what we choose the style for,' is purpose; some styles have specific features, such as colors, silhouettes, and patterns, that people consume as a fashion or represent a certain social identity and have some characteristic features. The Gal and Fairy styles are in the latter group and represent a 'way of life' for some people \cite{Hasegawa2015, Kyary2011}.

Icons representing 'way of life,' such as celebrities, have substantial power to make people behave in a certain manner; for instance, celebrities can make people buy products that they use. Choosing fashion styles is no exception, and some articles identified fashion icons for the Gal and the Fairy styles \cite{Shinmura2012, Hasegawa2015,Yoshimura2019}. We attempted to explain why the style trends in the streets showed different modes from the perspective of fashion icons and the media type that the icons utilized.

Figs. \ref{fig:fig15}, \ref{fig:fig16}, and \ref{fig:fig17} show the relationship between the style shares and 'magazine' trends for style icons. Previous works introduced the following Gal style icons: Namie Amuro for the mid-90s, SHIBUYA 109 for the late-00s, Ayumi Hamasaki for the early-00s, and Kumi Koda and Tsubasa Masuwaka for the late-00s \cite{Shinmura2012, Hasegawa2015}.

The first icon who boosted the Gal style was Namie Amuro, who debuted nationwide as a pop singer in 1992. Girls yearned to imitate her fashion style, and fashion magazines focused heavily on her as a style icon in the mid-90s. Simultaneously, as the 'magazine' trend for the style shows in Fig. \ref{fig:fig15}, many tabloid magazines were interested in this phenomenon and spread it as a new way of life for the youth. The first icon boosted the style in both streets because she appeared nationwide. However, the second icon that sustained the Gal style's upward trend is region specific: an iconic fashion mall named SHIBUYA 109 (pronounced Ichi-maru-kyū) in Shibuya. Many tenants in the mall sold Gal style products, which is why the Gal style is also known as 'maru-kyū fashion' \cite{Yoshimura2019}. This street-specific image created by the second icon might have influenced the third and fourth icons, Ayumi Hamasaki and Kumi Koda, who were nationwide pop singers in the early-00s and the mid-00s, because the style share increased during that time only in Shibuya.

The icons prompted the last increase in the Gal style's share in Shibuya, and the spike in the Fairy style's share in Harajuku have the same characteristics. Both Tsubasa Masuwaka for the Gal style in Shibuya and 'Kyary Pamyu Pamyu' for the Fairy style in Harajuku were active as exclusive reader models in street-based magazines at the beginning, and they created the booms in each street \cite{Shinmura2012}. However, the street-based magazines they belonged to were not included in the digital magazine archive in Oya Soichi Library; hence, the 'magazine' trends for their names missed their activities as exclusive reader models when the style shares showed an upward trend (Figs. \ref{fig:fig16} and \ref{fig:fig17}). Around 2010, they debuted as nationwide pop stars and frequently appeared in mass-circulation magazines and on television. Additionally, Tsubasa Masuwaka proposed a new style named 'Shibu-Hara' fashion, a combination of the styles in both Shibuya and Harajuku, and also referred to interactions with 'Kyary Pamyu Pamyu' on her social networking account. The Gal style's second peak in Harajuku and the Fairy style's spike in Shibuya in the early 2010s aligned with these icons' nationwide activities; this indicates the styles' cross-interactions driven by the icons (Figs. \ref{fig:fig16} and \ref{fig:fig17}).

\subsection{Summary and Discussion}
In previous studies, researchers pointed out that social contexts influence people's daily-life fashion. To our knowledge, our study is the first to discuss how much social contexts affect what people wear in a quantitative manner. Our findings in Section 5.1 suggest that our approach with CAT STREET can validate other fashion phenomena observed and discussed qualitatively by providing objective indices about daily-life fashion.

We explored the potential of our approach with CAT STREET to prompt research questions that can contribute to the expansion of fashion theories in Section 5.2. The first case in Section 5.2.1 prompted a new research question of determining the importance of the different roles that media play in fashion trends: the role of the 'mixed' pattern for creating and reporting new trends and the 'follow-up' effect to keep their momentum. To theorize how fashion trends are generated in more detail, our findings indicate that quantifying the trends from multiple digital archives is essential to test the research questions about the effect of the media's role, which has not been discussed thoroughly.

In previous studies, researchers analyzed fashion style trends at a street level independently. However, to analyze the recent, more complicated trend patterns, the second case in Section 5.2.2 suggests that it would be thought-provoking to focus on what the icons represent for people's social identities, rather than the style itself, and how they lead the trends. An example is the analysis of the types of media that icons use and the reach of that media. The viewpoints gained from our findings can be useful in tackling unsolved research questions such as how the styles interact with each other and how new styles are generated from this process. CAT STREET comprises images indicating what people wear in daily life. If we can determine people's social identities from “what they choose the fashion for” using the archive, CAT STREET can provide beneficial information to current research questions and play an important role in expanding fashion trend theories.

\section{Conclusions}
In this paper, we shed light on research questions about daily-life fashion trends with multiple digital archives using computer science methodologies. To quantify daily-life fashion trends, we built CAT STREET, which comprises fashion images illustrating what people wore in their daily lives in the period 1970–-2017, along with street-level geographical information. With DL, we demonstrated that the rules-of-thumb for the fashion trend phenomena, namely how economic conditions affect fashion style shares, how fashion styles emerge in the street, and the relationship between fashion styles and magazine trends, can be quantitatively validated using CAT STREET. Through empirical analyses to corroborate the rules-of-thumb, we demonstrated the potential of our database and approach to promote the understanding of our societies and cultures.

Our work is not without limitations. We used the fashion style categories of FashionStyle14 \cite{takagi2017makes}, which was defined by fashion experts and is considered to represent modern fashion. However, the definition does not cover all contemporary fashion styles and their substyles in a mutually exclusive and collectively exhaustive manner. Defining fashion styles is a complicated task because some fashion styles emerge from consumers, and suppliers define others. We must refine the definition of fashion styles to capture daily-life fashion trends more accurately.

Furthermore, prior to building CAT STREET, only printed photos were available for the period 1970–2009. Consequently, the numbers of images for these decades are not equally distributed because only those images from printed photos that have already undergone digitization are currently present in the database. The remainder of the printed photos will be digitized and their corresponding images added to the database in future work.

CAT STREET and our approach that combined multiple digital archives and methodologies in computer science to quantify fashion trends helped us explore undemonstrated research questions. For future studies that apply other quantitative analytical methods, such as unsupervised clustering, to extract fashion styles and social identities embedded in consumers' daily lives, CAT STREET will play a key role to find new standpoints and expand the boundaries of fashion studies.

\bibliographystyle{unsrt}

\end{document}